\documentclass[conference]{IEEEtran}
\IEEEoverridecommandlockouts
\usepackage{amsmath,amssymb,amsfonts}
\usepackage{booktabs} 
\usepackage{enumitem}
\usepackage{makecell} 
\usepackage{soul}
\usepackage{threeparttable}
\usepackage{multirow}
\usepackage{tabularx}
\usepackage{algorithmic}
\usepackage{algorithm} 
\usepackage{graphicx}
\usepackage{textcomp}
\usepackage{xcolor}
\usepackage{cite}
\usepackage{hyperref}
\usepackage{cleveref}
\usepackage{enumitem} 
\usepackage{etoolbox} 
\usepackage{epstopdf}

\setlist[enumerate]{leftmargin=*, nosep}
\setlist[itemize]{leftmargin=*, nosep}
\patchcmd{\section}{\centering}{\centering\small}{}{} 
\IEEEaftertitletext{\vspace{-2.0em}}

\newcommand{\vect}[1]{\boldsymbol{#1}}

\renewcommand{\baselinestretch}{0.96} 

\def\BibTeX{{\rm B\kern-.05em{\sc i\kern-.025em b}\kern-.08em
    T\kern-.1667em\lower.7ex\hbox{E}\kern-.125emX}}
    
\begin{document}

\title{Communication Outage-Resistant UUV State Estimation: A Variational History Distillation Approach
\thanks{This research was partially funded by Postgraduate Research Scholarship (PGRS) at Xi’an Jiaotong-Liverpool University (FOS2312JBD01), Suzhou Municipal Key Laboratory Broadband Wireless Access Technology (BWAT) and JITRI Supervision Support Fund (JSF10120220008) of XJTLU-JITRI Academy.}
\thanks{\copyright~2026 IEEE. Personal use of this material is permitted.  Permission from IEEE must be obtained for all other uses, in any current or future media, including reprinting/republishing this material for advertising or promotional purposes, creating new collective works, for resale or redistribution to servers or lists, or reuse of any copyrighted component of this work in other works.}
}

\author{
    \IEEEauthorblockN{
        Shuyue Li\textsuperscript{1,2,3}, 
        Miguel López-Benítez\textsuperscript{4,5}, 
        Eng Gee Lim\textsuperscript{1}, 
        Fei Ma\textsuperscript{6}, 
        Qian Dong\textsuperscript{1}, \\
        Mengze Cao\textsuperscript{3,7}, 
        Limin Yu\textsuperscript{1,*} and 
        Xiaohui Qin\textsuperscript{3,*}
    }
    \vspace{0.3em} 
    \IEEEauthorblockA{
        \footnotesize 
        \textsuperscript{1}School of Advanced Technology, Xi'an Jiaotong-Liverpool University, Suzhou, China \\
        \textsuperscript{2}School of Engineering, University of Liverpool, Liverpool, UK \\
        \textsuperscript{3}Jiangsu JITRI Tsingunited Intelligent Control Technology Co., Ltd., Wuxi, China \\
        \textsuperscript{4}School of Computer Science and Informatics, University of Liverpool, Liverpool, UK \\
        \textsuperscript{5}ARIES Research Centre, Universidad Antonio de Nebrija, Madrid, Spain \\
        \textsuperscript{6}School of Mathematics and Physics, Xi'an Jiaotong-Liverpool University (XJTLU), Suzhou, China \\
        \textsuperscript{7}College of Mechanical and Vehicle Engineering, Hunan University, Changsha, China \\
        \textsuperscript{*}Corresponding Authors: limin.yu@xjtlu.edu.cn, qinxiaohui@tsingunited.com
    }
   
}
\maketitle

\begin{abstract}
The reliable operation of Unmanned Underwater Vehicle (UUV) clusters is highly dependent on continuous acoustic communication. However, this communication method is highly susceptible to intermittent interruptions. When communication outages occur, standard state estimators such as the Unscented Kalman Filter (UKF) will be forced to make open-loop predictions. If the environment contains unmodeled dynamic factors, such as unknown ocean currents, this estimation error will grow rapidly, which may eventually lead to mission failure. To address this critical issue, this paper proposes a Variational History Distillation (VHD) approach. VHD regards trajectory prediction as an approximate Bayesian reasoning process, which links a standard motion model based on physics with a pattern extracted directly from the past trajectory of the UUV. This is achieved by synthesizing ``virtual measurements'' distilled from historical trajectories. Recognizing that the reliability of extrapolated historical trends degrades over extended prediction horizons, an adaptive confidence mechanism is introduced.  This mechanism allows the filter to gradually reduce the trust of virtual measurements as the communication outage time is extended. Extensive Monte Carlo simulations in a high-fidelity environment demonstrate that the proposed method achieves a 91\% reduction in prediction Root Mean Square Error (RMSE), reducing the error from approximately 170 m to 15 m during a 40-second communication outage. These results demonstrate that VHD can maintain robust state estimation performance even under complete communication loss.
\end{abstract}



\section{Introduction}
Unmanned Underwater Vehicles (UUVs) rely heavily on cooperative navigation (CN) to perform complex multi-vehicle tasks. However, due to the physical limitations of underwater acoustic communication, it is extremely difficult to maintain reliable collaborative navigation \cite{r8}. Compared with land radio networks, acoustic links are subject to many limitations, such as narrow bandwidth, high propagation delay, and serious multipath effects \cite{r8}. Given the challenges of maintaining continuous high-frequency data exchange in such environments, UUV networks are highly susceptible to prolonged communication outages.

When the communication is interrupted, the traditional state estimator, such as the Unscented Kalman Filter (UKF) \cite{r12}, will be forced to rely entirely on the open-loop position extrapolation. This working mode is highly sensitive to the mismatch of the kinematic model. The small difference between the assumed motion model (e.g., the constant acceleration hypothesis) and the actual UUV dynamics will quickly accumulate. When considering external interference such as ocean currents and sensor inherent noise, the positioning error will grow unbounded \cite{r3}. For instance, our baseline evaluations indicate that a standard predictor can drift more than $170$ meters during a short 40-second communication outage. Such a large positioning error can easily exceed the reliable detection range of typical forward sonar \cite{r9}, resulting in a high risk of navigation vehicle separation and overall mission failure \cite{r23}.

The existing methods for dealing with these faults can be roughly divided into two categories, but both types of methods have significant shortcomings. Purely model-based filters become biased due to their inability to adapt to unmodeled forces. On the contrary, the basic data-driven interpolation (e.g., polynomial fitting) ignores the physical limits of the vehicle and becomes extremely unstable in a longer prediction window, and the Runge's phenomenon often occurs \cite{r13}. Although deep learning models such as Long Short-Term Memory (LSTM) provide a powerful alternative to trajectory prediction, they require a lot of computational resources. For light UUV platforms with strict power restrictions, it is usually impractical to run these large neural networks onboard \cite{r14, r22, r5}. Therefore, there is an urgent need for a pragmatic alternative: a low-cost algorithm that can not only combine the structural stability of the filter based on physics, but also extract data-driven insights from the recent trajectory history of the vehicle.

To address this challenge, we proposed a new estimation scheme, namely, \textbf{Variational History Distillation (VHD)}. VHD re-expresses the trajectory extrapolation problem during the outages as an approximate Bayesian inference problem, thus providing a method based on mathematical principles to correct the position extrapolation error. The main contributions are summarized as follows:

\begin{enumerate}

\item We propose a novel VHD framework that reformulates trajectory prediction under communication outages as an approximate Bayesian inference problem.

\item We introduce a virtual measurement synthesis mechanism that distills historical trajectory information into a compact Gaussian representation compatible with recursive filtering.

\item We develop an adaptive confidence mechanism that ensures stable long-term prediction by dynamically balancing data-driven and model-based information.

\item Extensive simulations demonstrate that the proposed method significantly outperforms strong baselines, achieving an order-of-magnitude improvement in prediction accuracy compared to strong baselines in prediction accuracy under severe communication disruptions.

\end{enumerate}

The rest of the structure of this article is as follows: Section \ref{sec:related_work} briefly summarizes the existing literature on collaborative navigation and delay state estimation of UUV. Section \ref{sec:variational_framework} derived the mathematical basis of the VHD method and its adaptive confidence weighting. Section \ref{sec:simulation} introduces the simulation settings and our quantitative results. Section \ref{sec:discussion} discusses the operating advantages, limitations and parameter sensitivity of the proposed method, and finally summarizes the full text in section \ref{sec:conclusion}.

\section{Related Work}
\label{sec:related_work}
This section reviews the main methods of collaborative navigation and delay state estimation of UUV, and focuses on the specific technical gaps solved by the framework we proposed.

\subsection{Cooperative Navigation in the Undersea Domain}
CN is widely used to improve the positioning accuracy of multi-UUV networks. By broadcasting location and speed data in the cluster, the navigator can limit the inherent drift of its on-board Inertial Navigation Systems (INS) \cite{r15}. The existing architecture covers a simple leader-follower model to a distributed topology \cite{r16}. However, the performance of any collaborative navigation algorithm depends heavily on the stability of the acoustic link. Because acoustic channels often lose packets \cite{r8}, researchers are paying more and more attention to asynchronous and variable rate filtering. For example, a recent study has applied Variational Bayesian techniques to build an asynchronous filter that can tolerate delay measurement \cite{r18}.

Dealing with the sparseness of data in these networks continues to attract attention. Chen et al. \cite{r35} recently developed an adaptive finite-time tracking control method for heterogeneous Autonomous Underwater Vehicles (AUVs), while Hari et al. \cite{r36} used convex optimization to achieve robust attitude estimation under sparse measurement conditions. Although these recent studies have made significant progress in robust control and static optimization, they have not directly addressed the problem of state interpolation when communication is completely interrupted. The VHD framework is aimed at this specific problem by generating a virtual trajectory update when the external signal is completely lost.

\subsection{The Failure of Model-Based Filtering}
Traditional recursive estimators, such as Kalman filter (KF) \cite{r11} and its nonlinear variants such as the Extended Kalman Filter (EKF) and UKF \cite{r12}, rely on continuous prediction-update cycles. When the communication fails, the filter will miss the update step and adopt open-loop prediction by default. In this state, any mismatch between the hypothetical kinematic model (e.g., constant velocity) and the real motion of the UUV will lead to the rapid accumulation of estimated errors \cite{r17}.

Although advanced methods such as the Variational Bayesian Adaptive Kalman Filter (VBAKF) can dynamically adapt to unknown noise covariances \cite{r18}, they still need to input measurement data to trigger adaptation. In the case of a complete communication outage, these adaptive mechanisms will basically stop working. In order to overcome this limitation, our method synthesizes a predicted ``virtual measurement'' based on the historical trajectory of the vehicle, so that the filter can perform pseudo-updates even if the physical sensor does not provide any data.

\subsection{Data-Driven Methods}
The pure data-driven extrapolation method provides an alternative method for predicting the trajectory. Simple techniques such as polynomial interpolation \cite{r13} lack physical kinematic constraints, and due to the Runge's phenomenon \cite{r13}, they will quickly become unstable on a longer time scale. On the contrary, deep neural networks such as LSTMs show strong predictive power \cite{r14, r22}. However, they require a large amount of computing power and a huge set of training data, which makes it difficult for them to deploy on resource-limited embedded UUV hardware, especially in dynamic offshore environments.

\subsection{Variational Inference in State Estimation}
Variational Inference (VI) provides a computationally easy-to-process method to approximate complex a posterior distributions. By defining a simpler distribution of the family $q(\mathbf{s})$ to approximate the difficult-to-process real posteriori distribution $p(\mathbf{s}|\mathbf{z})$ (where $\mathbf{s}$ is a potential state, $\mathbf{z}$ is the observed value), VI transforms the integral problem into an optimization problem. The goal is to minimize the Kullback-Leibler (KL) divergence between $q(\mathbf{s})$ and $p(\mathbf{s}|\mathbf{z})$, which is equivalent to maximizing the Evidence Lower Bound (ELBO) \cite{r37}.

While VI is prevalent in Factor Graph Optimization (FGO) \cite{r24} and Simultaneous Localization and Mapping (SLAM) algorithms \cite{r19, r25, r31} for batch processing and noise parameter learning \cite{r18, r27}, our application of VI is fundamentally different. We adapt the concept of Variational Information Distillation (VID) \cite{r34} to compress recent trajectory history into a Gaussian proxy. This proxy serves as the aforementioned virtual measurement. Instead of merely tuning noise parameters, our framework uses VI to structurally compensate for data loss. Crucially, this maintains a recursive, lightweight $O(N)$ computational complexity, which is essential for real-time UUV operations.

\subsection{Synthesis: Bridging the Gap}
Overall, current delayed-state estimation strategies generally force a choice between model-based methods that ignore historical context during outages, and data-driven methods that ignore physical kinematic constraints (as detailed in our recent survey \cite{r5}). None of these methods explicitly addresses state estimation under complete communication blackout, where no external measurements are available. The VHD approach avoids this trade-off by using VI to fuse physical motion models with the corrective patterns extracted from recent historical data.

\section{A Variational Framework for Fusing Physical Priors with Historical Evidence}
\label{sec:variational_framework}
This section establishes the mathematical basis for the VHD framework. While the original Variational Information Distillation (VID) method was designed to transfer knowledge across neural networks \cite{r34}, we restructure its core mechanism to tackle state estimation problems. By applying VHD, we can extract underlying kinematic trends from past trajectory data and convert them into computationally manageable priors for Bayesian inference. We reformulate what might otherwise be an intuitive two-stage process into a principled, single-step approximate Bayesian inference procedure.

\subsection{Difference from Existing Paradigms}
Before explaining the derivation process in detail, it is crucial to distinguish between the VHD we proposed and the two related methods:

\begin{enumerate}
    \item \textbf{Compared with Deep Learning VID:} Unlike VID \cite{r34}, which maximizes mutual information to compress neural features, VHD extracts kinematic patterns from physical trajectories by minimizing KL-divergence.
    \item \textbf{Compared with Adaptive Filtering:} Unlike VBAKF \cite{r18}, which requires measurements to tune noise parameters, VHD operates during total outages by synthesizing virtual measurements ($\vect{z}^*$) from historical data.
\end{enumerate}

Although the proposed framework is inspired by variational inference principles, it should be interpreted as an information projection onto a constrained posterior family reachable via Kalman updates, rather than a full ELBO-based variational optimization.

\subsection{Probabilistic Problem Formulation}
We consider a \textbf{target UUV} operating in a 2D plane. Its state vector $\vect{s}_k$ at a discrete time step $k$ is defined as a six-dimensional vector containing absolute position ($p$), velocity ($v$), and acceleration ($a$) components:
\begin{equation}
\vect{s}_k = [p_{x,k}, v_{x,k}, a_{x,k}, p_{y,k}, v_{y,k}, a_{y,k}]^\mathrm{T}
\end{equation}
The UUV's motion can be described by a discrete-time linear stochastic system \cite{r3}:
\begin{equation}
\vect{s}_k = \vect{F} \vect{s}_{k-1} + \vect{w}_{k-1}
\end{equation}
where $\vect{F}$ is the state transition matrix based on a Constant Acceleration (CA) model, and $\vect{w}_{k-1}$ is the process noise, assumed to be zero-mean Gaussian white noise with covariance matrix $\vect{Q}$.

We assume that the observer's dynamic nature is not considered in this formulation, and the state variables represent absolute values in the global frame, not relative values.

The core problem is as follows: assume at time $t$, an observing UUV loses communication with a target UUV. The observer possesses a set of the target's historical state estimates and their covariances, denoted as the history set $\mathcal{H}_t = \{\vect{s}_{t-N|t-N}, \cdots, \vect{s}_{t|t}\}$, where $N$ denotes the length of the historical window. The communication outage will last for $T$ time steps. The task is to compute the optimal posterior probability distribution of the target's state at the future time $t+T$, $p(\vect{s}_{t+T} \mid \mathcal{H}_t)$, using only the information in $\mathcal{H}_t$.

\subsection{An Information-Theoretic View of State Correction}
To solve this, we model the update using a standard Bayesian approach that depends on two main distributions. 

\textbf{Physics-Based Prior $p(\vect{s}_{t+T})$:} When communication drops at time $t$, the system relies on the last known state $\vect{s}_{t|t}$ and its covariance $\vect{P}_{t|t}$. From this point, we run a $T$-step open-loop prediction via the CA model. The resulting Gaussian distribution acts as our prior:
\begin{equation}
\label{eq:prior_dist_1}
p(\vect{s}_{t +T}) = \mathcal{N}(\vect{s}_{t+T}; \vect{s}_{t+T|t}, \vect{P}_{t+T|t})
\end{equation}
where $\vect{s}_{t+T|t} = \vect{F}^T \vect{s}_{t|t}$, and $\vect{P}_{t+T|t}$ is computed by recursively applying $\vect{P}_{k|k-1} = \vect{F} \vect{P}_{k-1|k-1} \vect{F}^T + \vect{Q}$. Because this open-loop projection is driven entirely by the CA model, the state error accumulates continuously. Without new acoustic updates to bound this drift, the covariance matrix $\vect{P}_{t+T|t}$ inflates substantially as the communication outage $T$ lengthens.

\textbf{History-Based Target Distribution $q(\vect{s}_{t+T})$}: The historical trajectory set $\mathcal{H}_t$ captures the actual maneuvering characteristics of the UUV prior to communication loss. To extract this kinematic baseline, we perform a batch least-squares polynomial fit over the data points in $\mathcal{H}_t$, yielding a smoothed state estimate $\hat{\vect{s}}_t$. We use polynomial regression because it has the high efficiency and numerical robustness of closed solutions. For windows with a length of $N$, it only takes $O(N)$ operations. It is worth noting that the framework can be easily compatible with other regressors, such as spline functions or Gaussian processes. After obtaining $\hat{\vect{s}}_t$, we project it forward by $T$ steps through the CA model to define the mean vector $\vect{\mu}_q = \vect{F}^T \hat{\vect{s}}_t$. Then, fit the residual calculation matrix $\vect{\Sigma}_q$ according to the experience of historical data. Therefore, $q(\vect{s}_{t+T}) = \mathcal{N}(\vect{s}_{t+T}; \vect{\mu}_q, \vect{\Sigma}_q)$ can be used as a data-driven reference, and the estimated value will be anchored on the observed historical trajectory, instead of relying on pure navigation calculation.
\subsection{Deriving the Virtual Measurement}
Since the exact analytical computation of the posterior distribution often exceeds the on-board processing capabilities of standard UUVs, we employ variational inference to construct an alternative distribution $p'(\vect{s}_{t+T})$, so that it matches the data-driven target distribution $q(\vect{s}_{ t+T})$ \cite{r18}. The structural differences between these distributions are measured by KL-divergence \cite{r28}. For the Gaussian density distribution, minimizing the divergence is mathematically simplified into precise moment matching, thus avoiding a large amount of overhead caused by numerical integration. Practically, this closed-form property ensures the optimization process remains computationally lightweight, an absolute prerequisite for real-time execution on the resource-constrained embedded hardware typical of lightweight UUVs.

We define the tractable family of distributions $p'$ as the set of all possible posteriors that can be obtained by performing a single, standard Kalman measurement update on the prior distribution $p$. This update process is parameterized by a virtual measurement $\vect{z}$ and its associated noise covariance $\vect{R}^*$. The problem is thus transformed into an optimization problem: to find the optimal virtual measurement $\vect{z}^*$ that minimizes the KL-divergence between the resulting posterior $p'(\vect{z})$ and the target distribution $q$ \cite{r37}:
\begin{equation}
\vect{z}^* = \underset{\vect{z}}{\arg\min} \, \text{KL}(p'(\vect{z}) \| q)
\end{equation}
where the posterior $p'(\vect{z}) = \mathcal{N}(\vect{\mu}_{p'}, \vect{\Sigma}_{p'})$ has its mean and covariance given by the standard Kalman update equations \cite{r11}:
\begin{equation}
\vect{K} = \vect{P}_{t+T|t}\vect{H}^T(\vect{H}\vect{P}_{t+T|t}\vect{H}^T + \vect{R}^*)^{-1}
\label{eq1}
\end{equation}
\begin{equation}
\vect{\mu}_{p'} = \vect{s}_{t+T|t} + \vect{K}(\vect{z} - \vect{H}\vect{s}_{t+T|t})
\label{eq2}
\end{equation}
\begin{equation}
\vect{\Sigma}_{p'} = (\vect{I} - \vect{K}\vect{H})\vect{P}_{t+T|t}
\label{eq3}
\end{equation}
Here, $\vect{H}$ is the measurement matrix that extracts the position components from the state vector. The optimal $\vect{z}^*$ is not sourced from any physical sensor. Instead, it is the product of an information-theoretic ``distillation'' and ``compression'' of the high-dimensional, complex historical trajectory information. It condenses the holistic motion trend embedded in the historical data into a standard, low-dimensional information vector that can be directly assimilated by a Bayesian filter.

\subsection{An Information Geometry Perspective}
From the perspective of information geometry \cite{r38}, we can understand this optimization mechanism more deeply. By treating the probability distribution set as a statistical manifold, the KL-divergence can be used as a measure of the asymmetric distance between states. Therefore, the proposed update step is essentially an information projection.

We can map the components of the fusion strategy to the following geometric spaces:

\begin{itemize}
\item  All Gaussian distributions constitute the underlying statistical manifold.
\item The physics-based prior, $p$, establishes our initial coordinate on this manifold.
\item The historical target value, $q$, represents the ideal goal point we want to achieve.
\item A subset of all possible posterior distribution $p'(\vect{z})$ can be generated by a single Kalman update starting from $p$, forming a restricted sub-manifold.
\end{itemize}

Therefore, the goal of minimizing $\text{KL}(p'(\vect{z}) \| q)$ is converted into a geometric projection of the target $q$ to a sub-manifold that can achieve a posterior distribution. The optimal solution $p'(\vect{z}^*)$ obtained from this is the closest reach state measured by KL-divergence. This explanation confirms that our information fusion strategy does not rely on empirical heuristic methods, but on mathematically rigorous projections in the distribution space.

\subsection{Adaptive Confidence for Long-Term Prediction}
\label{sec:adaptive_confidence}
The operating utility of the refined virtual measurement $\vect{z}^{\ast}$ is fundamentally limited by the time span of the predicted time domain $T$. Although methods that rely on historical data fitting, such as polynomial extrapolation, can generate high-precision trajectory approximations immediately after sensor failure, this reliability is inherently short-lived. As the look-ahead interval $T$ grows, the coupling between past patterns and future states inevitably weakens. Consequently, any small fitting errors or deviations of the true motion from the recent historical pattern will be amplified over time, potentially leading to unstable or divergent state estimates if the filter continues to place high confidence in an increasingly inaccurate $\vect{z}^{\ast}$.

To address this fundamental limitation of long-term extrapolation, we introduce an adaptive confidence mechanism. This mechanism enables a smooth transition from data-driven correction to model-driven prediction. The core idea is to dynamically adjust the confidence placed in the virtual measurement based on the time elapsed since the communication outage began. We achieve this by making the virtual measurement noise covariance $\vect{R}^{\ast}$ time-varying.

Let $\Delta t = k \cdot dt - t_{\text{outage\_start}}$ be the time elapsed since the outage started at time step $k$. We define the dynamic virtual measurement noise covariance $\vect{R}_{k}^{\ast}$ at time step $k$ as:

\begin{equation}
\vect{R}_{k}^{\ast} = \vect{R}_{\text{base}} \cdot (1 + \alpha \cdot (\Delta t)^{p})
\end{equation}

Where:
\begin{itemize}
\item $\vect{R}_{\text{base}}$ is the base noise covariance, representing the high confidence (small uncertainty) we have in the virtual measurement immediately after the outage ($\Delta t = 0$). In practice, this is usually given a smaller value, such as $\text{diag}\{0.5, 0.5\}$.
\item $\alpha$ is the attenuation factor, which determines the sensitivity of uncertainty growth relative to the interruption duration.
\item $p$ is used as a growth index (usually $p=2$ for quadratic growth) to control the speed of confidence envelope expansion.
\end{itemize}

The proposed formula aims to capture two key stages of the state estimation of unmanned underwater navigation vehicles during the delay period:
\begin{enumerate}
    \item \textbf{Short-term Fidelity:} For near-instantaneous interruption ($\Delta t \approx 0$), we have $\vect{R}_{k}^{\ast} \approx \vect{R}_{\text{base}}$. This ensures that the estimator is highly dependent on historical fitting, which is still very accurate in the short term.
    \item \textbf{Dynamic Weight Transition:} With the increase of the duration $\Delta t$, the gradual expansion of $\vect{R}_{k}^{\ast}$ naturally reduces the weight of the virtual measurement $\vect{z}^{\ast}$ in the Kalman update (Eq. \ref{eq1}--\ref{eq3}). The mechanism shifts the fusion priority to the physics-based a priori prediction $p(\vect{s}_{t+T})$, effectively reducing the risk of external divergence.
\end{enumerate}

By adjusting $\vect{R}_{k}^{\ast}$ in real time, the filter realizes the intelligent trade-off between the initially accurate (but increasingly volatile) historical trajectory and the inherently stable (but may be biased) physical motion model. This adaptive balance is essential for maintaining robust UUV positioning throughout communication outages. In each discrete step $k$, use the instantaneous value of $\vect{R}_{k}^{\ast}$ to recursively calculate the updated equation (Eq. \ref{eq1}--\ref{eq3}).

\section{Validation in High-Fidelity Simulation}
\label{sec:simulation}
In order to evaluate the performance of our proposed framework under challenging and authentic conditions, we have designed a high-fidelity simulation environment in MATLAB. The scene is carefully designed to expose the shortcomings of traditional model-based estimates by introducing significant, unmodeled dynamic factors that reflect real ocean conditions.

\subsection{High-Fidelity Simulation Environment}
In order to bridge the gap between theoretical modeling and real-world deployment, our simulation incorporates two key real-world constraints:
\begin{itemize}
    \item \textbf{Uncompensated Ocean Currents:} Introduce a constant velocity vector in the real dynamics model to simulate environmental disturbance. The vector is strictly unobservable to all filter algorithms and can be used as a continuous external disturbance to test the robustness of the estimator. Its size ($\sim 1$~m/s) is calibrated to match the empirical data from coastal and port environments \cite{r28}.
    \item \textbf{Non-stationary Sensor Stochasticity:} We abandoned the simple Gaussian hypothesis and achieved high-fidelity noise characteristics for the Inertial Measurement Unit (IMU) accelerometer. The model takes into account the time-varying deviation through the random walking process, effectively capturing the inherent drift characteristics of low-cost MEMS sensors \cite{r29} commonly used in marine robots.
\end{itemize}
The reference trajectory is specially designed for a difficult maneuver stage, including a continuous high-acceleration turn before the communication is interrupted. The trajectory aims to break the CA hypothesis, so as to test the performance of the motion model under extreme kinematic conditions. To evaluate the estimators' dead-reckoning performance, a $40$-second communication outage was imposed. Furthermore, to ensure the statistical reliability of the performance metrics and mitigate the impact of stochastic noise, the performance improvement is consistent across all Monte Carlo trials, indicating strong robustness. The core simulation parameters are detailed in Table \ref{tab:settings}.

\begin{table}[htbp]
\caption{Simulation Parameter Settings}
\begin{center}
\begin{tabularx}{\columnwidth}{|l|X|}
\hline
\textbf{Parameter} & \textbf{Value/Description} \\
\hline
Ocean Current & 1.0 m/s (Unmodeled, Constant) \\
\hline
IMU Sensor Noise & White Noise + Random Walk Bias \\
\hline
Communication Outage & 40 seconds \\
\hline
History Window ($N$) & 50 seconds \\
\hline
Monte Carlo Runs & 100 \\
\hline
Base Covariance ($\vect{R}_{\text{base}}$) & $\text{diag}\{0.5, 0.5\}$ \\
\hline
Decay Rate ($\alpha$) & 0.01 \\
\hline
Growth Exponent ($p$) & 2 \\
\hline
\end{tabularx}
\label{tab:settings}
\end{center}
\end{table}

The simulation procedure for the proposed VHD algorithm during the outage is outlined in Algorithm \ref{alg:vhd}.

\begin{algorithm}
\caption{Simulation of VHD during Outage}
\begin{algorithmic}[1]
\REQUIRE History set $\mathcal{H}_t$, State $\vect{s}_{t|t}$, Outage duration $T$
\STATE \textbf{Initialization:} $\vect{P}_{t|t} \leftarrow$ Standard UKF Covariance
\STATE \textbf{Virtual Measurement Generation:} Fit polynomial $f_{poly}$ to $\mathcal{H}_t$
\FOR{$k = 1$ to $T$}
    \STATE $\Delta t \leftarrow k \cdot dt$
    \STATE \mbox{\textbf{Step 1:}} Predict state $\hat{\vect{s}}_{k|k-1}$ and $\vect{P}_{k|k-1}$ via CA model
    \STATE \mbox{\textbf{Step 2:}} Generate virtual measurement: \\
    \hspace{\algorithmicindent} $\vect{z}^*_k = f_{poly}(t_{curr})$
    \STATE \mbox{\textbf{Step 3:}} Calculate measurement noise covariance: \\
    \hspace{\algorithmicindent} $\vect{R}^*_k = \vect{R}_{\text{base}} \cdot (1 + \alpha \cdot (\Delta t)^p)$
    \STATE \mbox{\textbf{Step 4:}} Apply Variational Update (Eq. \ref{eq1}--\ref{eq3}) using $\vect{z}^*_k, \vect{R}^*_k$
\ENDFOR
\RETURN Trajectory $\{\hat{\vect{s}}_{t+1}, \dots, \hat{\vect{s}}_{t+T}\}$
\end{algorithmic}
\label{alg:vhd}
\end{algorithm}

Parameter Tuning: The key hyperparameters for the VHD algorithm, namely the base virtual measurement noise $\vect{R}_{\text{base}}$ and the confidence decay rate $\alpha$, were set based on the principles outlined in Section \ref{sec:adaptive_confidence}. $\vect{R}_{\text{base}}$ was chosen to be small ($\text{diag}\{0.5, 0.5\}$) reflecting the high accuracy of short-term polynomial extrapolation. The decay rate $\alpha$ ($0.01$) and exponent $p$ ($2$) were tuned empirically to achieve a balance where the filter trusts the virtual measurement strongly initially but rapidly reduces this trust as the outage progresses, preventing divergence while still leveraging historical trends. The history window $N$ was fixed at $50\text{s}$ to ensure a stable estimation baseline.

\subsection{Results and Analysis}
\begin{figure}[t]
    \centering
   
    \centerline{\includegraphics[width=0.83\columnwidth]{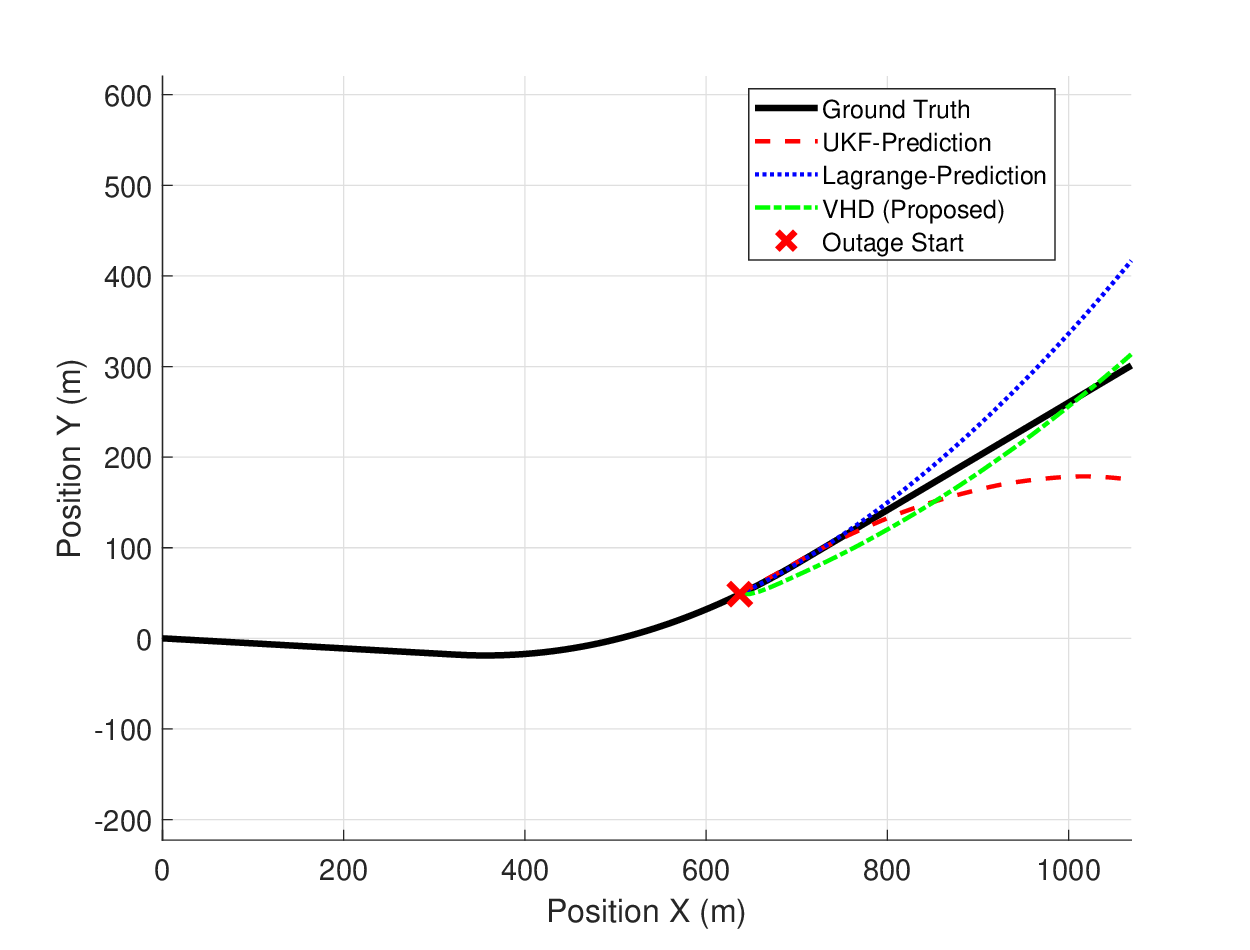}}
    \vspace{-1em} 
    \caption{Trajectory prediction comparison during the outage from a single run}
    \label{fig1}
\end{figure}

\begin{figure}[t]
    \centering 
   
    \centerline{\includegraphics[width=0.83\columnwidth]{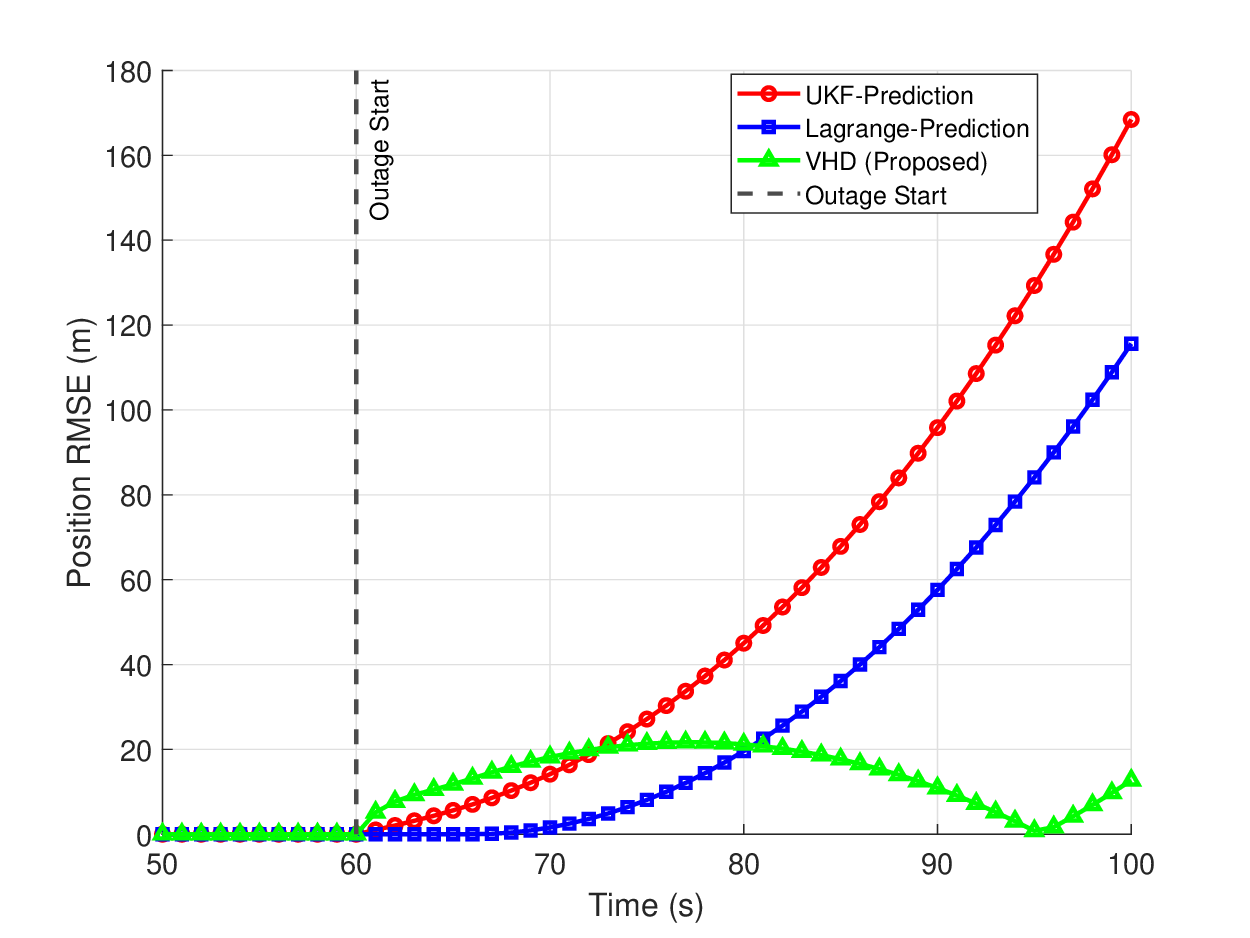}}
    \vspace{-1em} 
    \caption{Mean Position RMSE across 100 Monte Carlo realizations}
    \label{fig2}
\end{figure}

As shown in Fig.~\ref{fig1}, Fig.~\ref{fig2}, and in Table~\ref{tab_results}, the experimental results show that our proposed VHD framework is significantly better than the traditional method under unmodeled disturbance. It should be noted that because the motion model remains linear in this specific scenario, the function of the EKF is exactly the same as that of the standard KF, so it is not included in the comparison.

\textbf{Baseline Limitations:} The quantitative data in Table~\ref{tab_results} shows that the UKF model has a catastrophic failure, and the position error has soared to nearly $170$ meters. This deviation is mainly caused by unmodeled ocean currents and IMU deviations; the rigid internal model of the UKF model cannot compensate for these external forces, resulting in continuous drift from the real value. As shown in Fig.~\ref{fig1}, this is manifested in the continuous under-steering during the maneuvering of the underwater navigation vehicle. On the contrary, although the Lagrange predictor initially showed a certain stability, it will eventually produce unstable oscillations. This instability highlights the inherent risk of high-order polynomial extrapolation on a longer-term scale, which usually leads to excessive steering in the curve movement. These results highlight how vulnerable it is to rely only on fixed models or simple data-based extrapolation in dynamic underwater environments.

\textbf{VHD Performance and Mechanism:} In comparison, the VHD framework leverages an adaptive confidence mechanism to maintain stable and bounded errors throughout acoustic interruptions. As shown in Table~\ref{tab_results}, VHD has achieved a 91.1\% reduction in RMSE and limited the maximum error to about $15$ meters. The success of this method comes from its hybrid architecture, in which ``virtual measurement'' extracts potential motion patterns from the historical trajectory, thus implicitly capturing the influence of current and deviation, so that there is no need for explicit physical modeling. Adaptive logic prioritizes this correction signal in the short term, while controlling the growth of long-term uncertainty, and effectively linking state estimation with the real trajectory (Fig.~\ref{fig1}).

Regarding the error curve in Fig.~\ref{fig2}, the intermittent decline close to zero is not a sign of periodic convergence, but a physical by-product of the ``drift-pull'' interaction. Environmental disturbances force physics-based predictions to drift, while virtual measurements pull the estimated value back to the actual path. This correction cycle makes the predicted trajectory and the real trajectory intersect, and the error at the intersection is close to zero, thus effectively limiting the overall error range. 

\begin{table}[htbp]
\caption{Quantitative Performance \& Stability Comparison}
\centering 
\scriptsize 
\setlength{\tabcolsep}{6pt} 
\begin{tabular}{|l|c|c|c|}
\hline
\textbf{Algorithm} & \textbf{Stability} & \textbf{RMSE (m)} & \textbf{Reduct. (\%)} \\
\hline
UKF-Prediction & Diverges & $\approx 170$~m & Baseline \\
\hline
Lagrange-Prediction & Diverges & $\approx 120$~m & $\approx 29.4\%$ \\
\hline
\textbf{VHD (Proposed)} & \textbf{Bounded} & \textbf{$\approx 15$~m} & \textbf{$\approx 91.1\%$} \\
\hline
\end{tabular}
\label{tab_results} 
\end{table}

\section{Discussion}
\label{sec:discussion}
Our numerical experimental results emphasize the necessity of robust information fusion when the main communication link fails. In order to clarify the practical application value of the VHD framework, we analyze its performance characteristics and current limitations from the perspective of theoretical consistency and empirical observation.

\subsection{Key Strengths of the VHD Approach}
\begin{itemize}[leftmargin=*, nosep]
    \item \textbf{Error convergence during long-term communication outages:} Unlike model-centered estimates (e.g., UKF) or pure data-driven models that usually have divergent drifts during communication outages, VHD effectively anchors estimated errors. In our 40-second communication outage benchmark, RMSE was reduced by 91\% compared with the baseline method.
    
    \item \textbf{Adaptability to unstructured disturbances:} A significant advantage of VHD is that it can absorb unmodeled influencing factors, such as continuous ocean currents or sensor deviations, without explicit environmental templates. By refining the potential motion trend into virtual measurements, the filter can compensate for nonlinear factors that are difficult to parameterize.
    
    \item \textbf{Feasibility of on-board calculation:} We design VHD as an $O(N)$ recursive process, in which $N$ is the backtracking window. By compressing high-dimensional historical data into a single location-based update, the overhead of each step is equivalent to that of the standard Kalman filter. This efficiency is crucial for UUVs with limited embedded processing capacity, which can avoid the common delay problems of deep learning or complex factor diagrams.
\end{itemize}

\subsection{Current Limitations and Constraints}
\begin{itemize}[leftmargin=*, nosep]
    \item \textbf{Dimensional and kinematic hypothesis:} At present, our verification is limited to decoupling motion on the two-dimensional plane. Although this is sufficient to meet the needs of plane measurement tasks, it has not considered the complete 6-DOF spatial coupling unique to agile underwater maneuvers in a three-dimensional environment, such as sudden pitch or rolling fluctuations.
    
    \item \textbf{Extrapolation Decay over Time:} The reliability of any extrapolation based on historical data (including our polynomial fitting) will inevitably decrease with the increase of the prediction gap ($T$). Although our adaptive confidence mechanism will readjust the weight of the filter to make it closer to the physical model to prevent catastrophic failures, in the event of a very long-lasting communication outage, the information value of ``virtual measurement'' will inevitably reach the critical point of diminishing returns.
    
    \item \textbf{Dependency on Initialization Data:} VHD is essentially retrospective; it requires a historical data set filled with data ($\mathcal{H}_t$) to run. Therefore, if a communication failure occurs immediately after deployment and a sufficient trajectory buffer has not been established, the framework cannot provide corrections.
    
\end{itemize}

\subsection{Maneuver-Aware Parameter Adaptation}
The performance of VHD is sensitive to the attenuation rate ($\alpha$) and the growth index ($p$), and both must be consistent with the kinematic behavior of UUV.

\begin{itemize}[leftmargin=*, nosep]
    \item \textbf{Sensitivity to Trajectory Dynamics:} These parameters control the speed at which the system gives up trust in historical patterns. For steady-state motion, the lower $\alpha$ value allows the filter to take advantage of the long-term trend. However, polynomial fitting may produce artifacts during high-acceleration maneuvers (e.g., fast S-turns). In this case, a larger $\alpha$ value is needed to prioritize physics-based models and avoid being misled by outdated trends.
    
    \item \textbf{Future Direction: Self-Supervised Tuning:} In order to achieve complete autonomous deployment, we proposed a cross-verification scheme, in which the UUV divides its historical data before failure into training sets and verification sets. By testing the extrapolation accuracy on its own historical data, the navigator can dynamically optimize the parameters $\alpha$ and $p$ in real time. This will enable the distillation process to adaptively respond to the complexity of the current maneuvers.
    
\end{itemize}

\section{Conclusion}
\label{sec:conclusion}
In this study, we have proposed a Variational History Distillation (VHD) method, which bridges the gap between physically driven motion modeling and historical trajectory statistics. VHD extracts virtual measurement values by minimizing the KL-divergence and adopts an adaptive weighting scheme, which successfully alleviates the common rapid error accumulation problem during communication blackouts. The simulation results show that compared with the traditional model-based navigation method, this method can improve the prediction accuracy by more than 90\%.

Looking to the future, we plan to expand this architecture to six degrees of freedom (6-DOF) dynamics to support more complex underwater operations. Furthermore, we plan to validate the framework using semi-physical (hardware-in-the-loop) simulations, ensuring that VHD is robust enough for field trials in multi-UUV collaborative navigation scenarios.

\section*{Acknowledgment}
The authors thank Dr. Fanxin Wang (XJTLU) and Dr. Alessandro Varsi (UoL) for their insightful review feedback. Dr. Wang's advice clarified the mathematical formulation, while Dr. Varsi's recommendations strengthened the experimental validation. We also gratefully acknowledge IEEE OES for the OCEANS SPC grant supporting the first author's travel and registration. Additionally, the authors acknowledge the use of Gemini for language editing and technical proofreading during the manuscript preparation.

\begingroup
    \renewcommand{\baselinestretch}{0.94} 
    \selectfont
    \bibliographystyle{IEEEtran}
    \bibliography{ref}
\endgroup

\end{document}